\documentclass{article}
\usepackage{ijcai19}

\usepackage{times}
\usepackage{soul}
\usepackage{url}
\usepackage[hidelinks]{hyperref}
\usepackage[utf8]{inputenc}
\usepackage[small]{caption}
\usepackage{graphicx}
\usepackage{amsmath}
\usepackage{booktabs}
\usepackage{algorithm}
\usepackage{algorithmic}
\title{ASP-based Discovery of Semi-Markovian Causal Models under Weaker Assumptions}

\author{
Zhalama$^1$\footnote{Contact Author}\
\and
Jiji Zhang$^2$\and
Frederick Eberhardt$^{3}$\and
Wolfgang Mayer$^1$\And
Mark Junjie Li$^4$
\affiliations
$^1$University of South Australia\\
$^2$Lingnan University\\
$^3$California Institute of Technology\\
$^4$Shenzhen University
\emails
zhalama@mymail.unisa.edu.au,
jijizhang@ln.edu.hk,
fde@caltech.edu,
wolfgang.mayer@unisa.edu.au,
jj.li@szu.edu.cn
}

\usepackage{color} 
\usepackage{graphicx} 
\usepackage{url}
\usepackage{amsmath,amssymb} 
\usepackage{natbib} 
\usepackage{framed}
\usepackage{algorithm}
\usepackage{algorithmic} 
\usepackage{pdfpages}
\usepackage[toc,page]{appendix}

\newcommand{\given}{{ \; | \; }}

\newcommand{\dsep}{\perp}

\newcommand\independent{\protect\mathpalette{\protect\independenT}{\perp}} 
\def\independenT#1#2{\mathrel{\rlap{$#1#2$}\mkern2mu{#1#2}}}
\newcommand{\nindep}{{\hspace{1mm}\backslash \hspace{-3.5mm} \independent}}

\newcommand{\C}{{\bf C}}
\newcommand{\K}{{\bf{K}}}
\newcommand{\V}{{\bf{V}}}
\newcommand{\Z}{{\bf{Z}}}

\newcommand{\ra}{\rightarrow}
\newcommand{\la}{\leftarrow}
\newcommand{\lra}{\leftrightarrow}
\newcommand{\bia}{\leftrightarrow}

\usepackage{enumitem}
\newlist{steps}{enumerate}{1}
\setlist[steps, 1]{label = S \arabic*:}

\usepackage{amsthm}

\theoremstyle{definition}
\newtheorem{definition}{Definition}[]

\begin{document}

\maketitle

\begin{abstract}
  In recent years the possibility of relaxing the so-called Faithfulness assumption in automated causal discovery has been investigated. The investigation showed (1) that the Faithfulness assumption can be weakened in various ways that in an important sense preserve its power, and (2) that weakening of Faithfulness may help to speed up methods based on Answer Set Programming. However, this line of work has so far only considered the discovery of causal models without latent variables. In this paper, we study weakenings of Faithfulness for constraint-based discovery of semi-Markovian causal models, which accommodate the possibility of latent variables, and show that both (1) and (2) remain the case in this more realistic setting. 
\end{abstract}

\section{Introduction}

Causal inference is of great interest in many scientific areas, and automated discovery of causal structure from data is drawing increasingly more attention in the field of machine learning. One of the standard approaches to automated causal discovery, known as the constraint-based approach, seeks to infer from data statistical relations among a set of random variables, and translate those relations into constraints on the underlying causal structure so that features of the causal structure may be determined from the constraints \citep{sgs2000,pearl2000}. In this approach, the most commonly used constraints are in the form of conditional (in)dependence, which can serve as constraints on the causal structure due in the first place to the well known causal Markov assumption. The assumption states roughly that a causal structure, as represented by a directed acyclic graph (DAG), entails a certain set of conditional independence statements. With this assumption, a conditional dependency found in the data constrains the causal DAG.    


The causal Markov assumption is almost universally accepted by researchers on causal discovery. However, by itself the assumption is too weak to enable interesting causal inference \citep{Zhang2013}. It is therefore usually supplemented with an assumption known as Faithfulness, which states roughly that unless entailed by the causal structure according to the Markov assumption, no conditional independence relation should hold. With this assumption, conditional independence relations found in the data also constrain the causal DAG.         

Unlike the causal Markov assumption, the Faithfulness assumption is often regarded as questionable. The standard defense of the assumption is that violations of Faithfulness involve fine-tuning of parameters (such as two causal pathways balancing out exactly), which is very unlikely if we assume parameter values are somehow randomly chosen. However, parameter values may not be randomly chosen, especially in situations where balancing of multiple causal pathways may be part of the design. More importantly, even if the true distribution is faithful to the true causal structure, with finite data, ``apparent violations'' of faithfulness can result from errors in statistical tests, when a false hypothesis of conditional independence fails to be rejected. Such apparent violations of faithfulness cannot be reasonably assumed away \citep{Uhler2013} and will bring troubles to causal discovery that assumes Faithfulness \citep{meek96, RobinsEtal2003}. 

For these reasons, in recent years the possibility of relaxing the Faithfulness assumption has been investigated \citep{ramsey06, zs2008, Zhang2013, sz2014, RU2014, ForsterEtal2017}. This line of work made it clear that in the context of learning causal models with no latent variables, the Faithfulness assumption can be weakened or generalized in a number of ways while retaining its inferential power, because in theory these assumptions all reduce to the Faithfulness assumption when the latter happens to hold. 

On a more practical note, causal discovery algorithms have also been developed to fit some of these weaker assumptions, most notably the Conservative PC algorithm \citep{ramsey06} and the greedy permutation-based algorithms \citep{WangSYU17, Solus2018}. More systematically, \citet{Zhalama2017SATBasedCD}
implemented and compared a number of weakenings of Faithfulness in the flexible approach to causal discovery based on Answer Set Programming (ASP) \citep{hej2014}. Among other things, they found, rather surprisingly, that some weakenings significantly boost the time efficiency of ASP-based algorithms. Since the main drawback of the ASP-based approach lies with its feasibility, this finding is potentially consequential for the further development of this approach.        

However, neither the theoretical investigation nor the ASP-based practical exploration went beyond the limited (and unrealistic) context of learning causal models in the absence of latent confounding, also known as causal discovery with the assumption of \emph{causal sufficiency} \citep{sgs2000}. Since latent confounding is ubiquitous, it is a serious limitation to restrict the study to causally sufficient settings. And it is especially unsatisfactory from the perspective of the ASP-based approach, which boasts the potential to deal with a most general search space that accommodates the possibility of latent confounding and that of causal loops \citep{hhej2013}. 

In this paper, we make a step towards remedying this limitation by generalizing the aforementioned investigation in a setting where latent confounding is allowed (but not causal loops; we remark on a complication that will arise in the presence of causal loops in the end.) Since the investigation appeals to the ASP-based platform, we will follow previous work on this topic to use semi-Markovian causal models to represent causal structures with latent confounders. Among other things, we show that it remains the case that (1) the Faithfulness assumption can be weakened in various ways that in an important sense preserve its power, and (2) weakening of Faithfulness may help to speed up ASP-based methods.        

The remainder of the paper will proceed as follows. In Section \ref{sec: pre}, we introduce terminologies and describe the basic setup. In Section \ref{sec: weakeningDAG}, we review a few ways to relax the Faithfulness assumption that have been proposed in the context of causal discovery with causal sufficiency and have been proved to be {\it conservative} in a sense we will specify. Then, in Section \ref{sec: weakeningSMCM}, we discuss the complications that arise with semi-Markovian causal models, and establish generalizations of the results mentioned in Section \ref{sec: weakeningDAG}. This is followed by a discussion in Section \ref{sec: encoding} of how to implement the weaker assumptions in the ASP platform. Finally, we report some simulation results in Section \ref{sec: simulations} that demonstrate the speed-up mentioned above, and conclude in Section \ref{sec: conclusion}. 

\section{Preliminaries}
\label{sec: pre} 


In this paper, the general graphical representation of a causal structure is by way of a {\it mixed graph}. The kind of mixed graph we will use is a triple $(\mathbf{V},\mathbf{E}_1, \mathbf{E}_2)$, where $\mathbf{V}$ is a set
of vertices (each representing a random variable), $\mathbf{E}_1$ a set of directed edges ($\ra$) and $\mathbf{E}_2$ a set of bi-directed edges ($\bia$). In general, more than one edge is allowed between two vertices, but no edge is allowed between a vertex and itself. Two vertices are said to be {\it adjacent} if there is at least one edge between them. Given an edge $X \ra Y$, $X$ is called a {\it parent} of $Y$ and $Y$ a {\it child} of $X$. We also say the edge has a {\it tail} at $X$ and an {\it arrowhead} at $Y$. An edge $X \bia Y$ is said to have an arrowhead at both $X$ and $Y$. 
A {\it path} between $X$ and $Y$ consists of an ordered sequence of distinct vertices $\langle X=V_1, ..., V_n=Y \rangle$ and a sequence of edges $\langle E_1, ..., E_{n-1}\rangle$ such that for $1\leq i\leq n-1$, $E_i$ is an edge between $V_i$ and $V_{i+1}$.
Such a path is a {\it directed path} if for all $1\leq i\leq n-1$, $E_i$ is a directed edge from $V_i$ to $V_{i+1}$. $X$ is an {\it ancestor} of $Y$ and $Y$ an {\it descendant} of $X$, if either $X = Y$ or there is a directed path from $X$ to $Y$. A {\it directed cycle} occurs when two distinct vertices are ancestors of each other. 


If a mixed graph does not contain any directed cycle, we will call it a {\it semi-Markovian causal model} (SMCM), also known as an {\it acyclic directed mixed graph} (ADMG). Intuitively a directed edge in an SMCM represents a direct causal relationship, and a bi-directed edge represents the presence of latent confounding. A directed acyclic graph (DAG) is a special case where no bi-directed edge appears. A DAG can be thought of as representing a causal model over a {\it causally sufficient} set of random variables, which may be referred to as a Markovian causal model (MCM).     


The conditional independence statements entailed by a graph can be determined graphically by a separation criterion. 
One statement of this criterion is {\it m-separation}, which is a natural generalization of the celebrated {\it d-separation} criterion for DAGs \citep{Pearl88}. 
Given any path in a mixed graph $G$, a non-endpoint vertex $V$ on the path is said to be a {\it collider} on the path if both edges incident to $V$ on the path have an arrowhead at $V$. Otherwise it is said to be a {\it non-collider} on the path.

\theoremstyle{definition}
\begin{definition}[m-connection and m-separation]
 Given a mixed graph $G$ over $\mathbf{V}$ and $\Z \subseteq \V$, a path in $G$ is {\it m-connecting given $\Z$} if every non-collider on the path is not in $\mathbf{Z}$ and every collider on the path has a descendant in $\mathbf{Z}$.
 
For any distinct $X, Y\notin \Z$, $X$ and $Y$ are {\it m-separated by $\mathbf{Z}$} in $G$ (written as $X \dsep_G Y \given \Z$) if there is no path between $X$ and $Y$ that is m-connecting given $\mathbf{Z}$. Otherwise $X$ and $Y$ are said to be {\it m-connected by $\mathbf{Z}$}.

For any $\mathbf{X}, \mathbf{Y}, \mathbf{Z}\subseteq \mathbf{V}$ that are pairwise disjoint, $\mathbf{X}$ and $\mathbf{Y}$ are m-separated by $\mathbf{Z}$ in $G$ if every vertex in $\mathbf{X}$ and every vertex in $\mathbf{Y}$ are m-separated by $\mathbf{Z}$. 
\end{definition}

\begin{figure}[h]
\centerline{
{\includegraphics[width=0.35\textwidth]{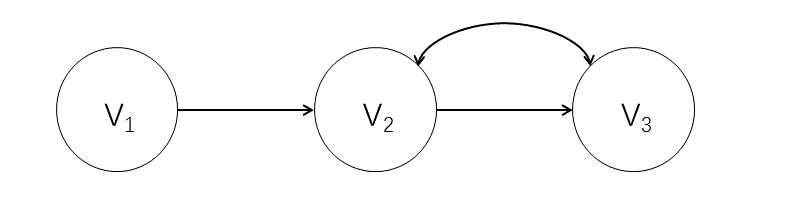}}
}

\caption{An inducing path between two non-adjacent vertices}
\label{fig: inducing}
\end{figure}

This definition obviously reduces to that of d-connection and d-separation in the case of DAGs. It is well known that in a DAG, two vertices are adjacent if and only if no set of other vertices d-separates them. The `only if' direction holds for SMCMs, but the `if' direction does not. For example, in the simple SMCM in Figure \ref{fig: inducing}, $V_1$ and $V_3$ are not adjacent, but neither the empty set nor the set $\{V_2\}$ m-separates them. This motivates the following definition.

\theoremstyle{definition}
\begin{definition}[inducing path]
A path between $X$ and $Y$ is an {\it inducing path} if every non-endpoint vertex on the path is a collider and also an ancestor of either $X$ or $Y$.
\end{definition}
For example, in Figure \ref{fig: inducing}, the path $V_1 \ra V_2 \bia V_3$ is an inducing path between $V_1$ and $V_3$. In general, two vertices in an SMCM are not m-separated by any set of other variables if and only if there is an inducing path between them \citep{Verma1993}. Note that an edge between two vertices constitutes an inducing path. Following \citet{RICHARDSON1997}, we call two vertices {\it virtually adjacent} if there is an inducing path between them. Adjacency entails virtual adjacency, but not vice versa.

\section{Faithfulness and its weakening for learning causal models without latent variables}
\label{sec: weakeningDAG}
We now review some proposals of weakening the Faithfulness assumption in the context of learning (acyclic) causal structures in the absence of latent confounding. In such a case, the target is a DAG over the given set of random variables $\mathbf{V}$, in which each edge represents a direct causal relation relative to $\mathbf{V}$ \citep{sgs2000}. Let $G$ denote the unknown true causal DAG over $\mathbf{V}$, and $P$ denote the true joint probability distribution over $\mathbf{V}$. The causal Markov assumption can be formulated as: \\
\\
{\bf Causal Markov assumption} For every pairwise disjoint $\mathbf{X}, \mathbf{Y}, \mathbf{Z}\subseteq \mathbf{V}$, if $\mathbf{X} \dsep_G \mathbf{Y} \given \Z$, then $\mathbf{X} \independent_{P} \mathbf{Y} \given \Z$. \\
\\
where `$\mathbf{X} \dsep_G \mathbf{Y} \given \Z$' means that $\mathbf{X}$ and $\mathbf{Y}$  are d-separated by $\mathbf{Z}$ in $G$, and `$\mathbf{X} \independent_{P} \mathbf{Y} \given \Z$' means that $\mathbf{X}$ and $\mathbf{Y}$ are independent conditional on $\mathbf{Z}$ according to $P$.   

The converse is the Faithfulness assumption: \\
\\
{\bf Causal Faithfulness assumption} For every pairwise disjoint $\mathbf{X}, \mathbf{Y}, \mathbf{Z}\subseteq \mathbf{V}$, if $\mathbf{X} \independent_P \mathbf{Y} \given \Z$, then $\mathbf{X} \dsep_{G} \mathbf{Y} \given \Z$. \\

As mentioned earlier, the Faithfulness assumption is regarded as much more questionable than the Markov assumption, and the literature has seen a number of proposals to relax it. In this paper, we focus on the following three.\footnote{Another two proposals are known as `Triangle-Faithfulness plus SGS-minimality' \citep{sz2014} and `P-minimality' \citep{Zhang2013}. It is not yet clear how to implement the latter in ASP, and the former did not seem to help much with ASP-based methods \citep{Zhalama2017SATBasedCD}.} \\
\\
{\bf Adjacency-faithfulness assumption} For every distinct $X, Y\in \mathbf{V}$, if $X$ and $Y$ are adjacent in $G$, then $X\nindep_P Y \given \Z$, for every $\Z \subseteq \V \setminus \{X, Y\}$. \\
\\
{\bf Number-of-Edges(NoE)-minimality assumption}: $G$ is NoE-minimal in the sense that no DAG with a smaller number of edges than $G$ satisfies the Markov assumption with $P$. \\
\\
{\bf Number-of-Independencies(NoI)-minimality assumption}: $G$ is NoI-minimal in the sense that no DAG that entails a greater number of conditional independence statements than $G$ does, satisfies the Markov assumption with $P$. \\

Under the Markov assumption, these assumptions are all weaker than the Faithfulness assumption. In words, Adjacency-faithfulness says that two variables that are adjacent in the causal structure are dependent given any conditioning set. It was first introduced in \citet{ramsey06} and motivated the CPC (conservative PC) algorithm. NoE-minimality says that the true causal structure has the least number of edges among all structures that satisfy the Markov assumption. It underlies the novel permutation-based algorithms that were developed recently \citep{RU2014, WangSYU17, Solus2018}. NoI-minimality says that the true causal structure entails the greatest number of conditional independence statements among all structures that satisfy the Markov assumption. In the ASP-based methods, the `hard-deps' conflict resolution scheme in \citet{hej2014} happened to implement this minimality constraint.

Theoretically these assumptions are particularly interesting because although they are weaker than Faithfulness (given the Markov assumption), they are in a sense strong enough to preserve the inferential power of Faithfulness. It has been shown that when Faithfulness happens to hold, all these weaker assumptions become equivalent to Faithfulness \citep{Zhalama2017SATBasedCD}. In other words, while they are weaker than Faithfulness and therefore still hold in many cases when Faithfulness does not, they rule out 
exactly the same causal graphs as Faithfulness does when the latter happens to be satisfied. We propose to call this kind of weakening {\it conservative}, for it retains the inferential power of Faithfulness whenever Faithfulness is applicable. The choice between a stronger assumption and a weaker one usually involves a trade-off between risk (of making a false assumption) and inferential power, but there is no such trade-off if the weakening is conservative. 


In addition to this theoretical virtue, both Adjacency-faithfulness and NoE-minimality, and especially Adjacency-faithfulness, have been shown to significantly improve the time efficiency of ASP-based causal discovery methods, without significant sacrifice in performance  \citep{Zhalama2017SATBasedCD}. We aim to extend these findings to the much more realistic setting where latent confounding may be present. 

\section{Weakening Faithfulness for learning semi-Markovian causal models}
\label{sec: weakeningSMCM}


When the set of observed variables $\V$ is not causally sufficient, which means that some variables in $\V$ share a common cause or confounder that is not observed, it is no longer appropriate to represent the causal structure in question with a DAG over $\V$. One option is to explicitly invoke latent variables in the representation and assume the underlying causal structure is properly represented by a DAG over $\V$ plus some latent variables $\mathbf{L}$. Another option is to suppress latent variables and use bi-directed edges to represent latent confounding. The use of SMCMs exemplifies the latter approach.\footnote{Another important example is the use of ancestral graph Markov models \citep{Richardson02ancestralgraph}, which we describe in the appendices.} 

As \citet{Verma1993} showed, for every DAG over $\V$ and set of latent variables $\mathbf{L}$, there is a unique projection into an SMCM over $\V$ that preserves both the causal relations among $\V$ and the entailed conditional independence relations among $\V$. Moreover, as \citet{Richardson03markovproperties} pointed out, the original causal DAG with latent variables and its projection into an SMCM are equivalent regarding the (nonparametric) identification of causal effects. These facts justify using SMCMs to represent causal structures with latent confounding.

So let us suppose the underlying causal structure over $\V$ is properly represented by an SMCM $G$ and let $P$ denote the true joint distribution over $\V$. In this setting, the causal Markov and Faithfulness assumptions can be formulated as before (in Section \ref{sec: weakeningDAG}), except that the separation criterion is now understood as the more general m-separation. Next we examine the proposals of weakening Faithfulness.

Regarding Adjacency-faithfulness, it is easy to see that it remains a logical consequence of Faithfulness. If two variables are adjacent in an SMCM, then given any set of other variables, the two are m-connected (any edge between them constitutes a m-connecting path). Thus, if Faithfulness holds, then they are not independent conditional on any set of other variables, exactly what is required by Adjacency-faithfulness. Since Adjacency-faithfulness does not entail Faithfulness in the case of DAGs and DAGs are special cases of SMCMs, Adjacency-faithfulness remains weaker than Faithfulness.

However, it is now too weak to be a conservative weakening of Faithfulness. Here is a very simple example. Suppose the true causal structure over three random variables is a simple causal chain $V_1\ra V_2\ra V_3$, and suppose the joint distribution is Markov and Faithful to this structure. So we have $V_1\independent V_3\given V_2$. Then the distribution is not Faithful to the structure in Figure \ref{fig: inducing}, because that structure does not entail that $V_1$ and $V_3$ are m-separated by $V_2$. Still, Adjacency-faithfulness is satisfied by the distribution and the structure in Figure \ref{fig: inducing}, for the only violation of Faithfulness occurs with regard to $V_1$ and $V_3$, which are not adjacent. Therefore, in this simple case where Faithfulness happens to hold, if we just assume Adjacency-faithfulness, we are not going to rule out the structure in Figure \ref{fig: inducing}, which would be ruled out if we assumed Faithfulness.

This simple example suggests that we should consider the following variation: \\
\\
{\bf V(irtual)-adjacency-faithfulness assumption}: 
For every distinct $X, Y\in \mathbf{V}$, if $X$ and $Y$ are virtually adjacent in $G$ (i.e., if there is an inducing path between $X$ and $Y$ in $G$), then $X\nindep_P Y \given \Z$, for every $\Z \subseteq \V \setminus \{X, Y\}$. \\

V-adjacency-faithfulness is obviously stronger than Adjacency-faithfulness, but we can prove that it remains weaker than Faithfulness. More importantly, it is strong enough to be a conservative weakening of Faithfulness.

How about NoE-minimality? Since more than one edge can appear between two vertices,  NoE-minimality (as it is formulated in Section \ref{sec: weakeningDAG}) is no longer a consequence of Faithfulness. To see this, just suppose the true structure over two random variables is simply $V_1\ra V_2$ together with $V_1\bia V_2$ (i.e., $V_1$ is a cause of $V_2$ but the relation is also confounded), and suppose the distribution is Markov and Faithful to this structure. NoE-minimality is violated here, for taking away either (but not both) of the two edges still results in a structure that satisfies the Markov assumption. 

So NoE-minimality is not a weakening of Faithfulness. Note that in the case of DAGs, minimization of the number of edges is equivalent to minimization of the number of adjacencies. If we replace the former with the latter, the above example is taken care of (for taking away the adjacency in that example will result in a structure that fails the Markov assumption). However, it is also easy to construct an example where an adjacency in an SMCM can be taken away without affecting the independence model \citep{Richardson02ancestralgraph}, so adjacency-minimality also fails to be a weakening of Faithfulness. The right generalization of NoE-minimality is unsurprisingly the following:    \\   
\\
{\bf V(irtual)-adjacency-minimality assumption}: $G$ is V-adjacency-minimal in the sense that no SMCM with a smaller number of virtual adjacencies than $G$ satisfies the Markov assumption with $P$. \\




 

Finally, since NoI-minimality is concerned with entailed conditional independence statements, it is straightforwardly generalized to the setting of SMCMs (just replace `DAG' with `SMCM' in the original formulation), and remains a conservative weakening of Faithfulness. Here then is the main result of this section (a proof of which is given in Appendix C). 

\newtheorem*{theorem*}{Theorem}
\begin{theorem*}
Given the causal Markov assumption, the V-adjacency-faithfulness assumption, V-adjacency-minimality assumption, and NoI-minimality assumptions are all conservative weakenings of the Faithfulness assumption, in the following sense: for each of the three assumptions {\bf AS},
\begin{enumerate}
    \item[(a)] {\bf AS} is entailed by, but does not entail, Faithfulness. 
    \item[(b)] For every joint probability distribution $P$ over $\V$, if there exists an SMCM that satisfies both Markov and Faithfulness assumptions with $P$, then for every SMCM $G$ that satisfies the Markov assumption with $P$, $G$ satisfies Faithfulness if and only if $G$ satisfies {\bf AS} with $P$.   
\end{enumerate}
\end{theorem*}


\section{ASP-based Causal Discovery of SMCMs}
\label{sec: encoding}

We instantiated causal discovery algorithms, which adopt V-adjacency-faithfulness and V-adjacency-minimality, using the framework of \citet{hej2014}. This framework offers a generic constraint-based causal discovery method based on Answer Set Programming (ASP). The logic is used to define Boolean atoms that represent the presence or absence of a directed or bi-directed edge in an SMCM. In addition, conditional independence/dependence statements (CI/CDs) obtained from tests on the input data are encoded in this logic. Finally, background assumptions, such as Markov and Faithfulness, are written as logical constraints enforcing a correspondence between the encoded test results and the underlying Boolean atoms (the edges of the SMCM). Solutions, which are truth-value assignments to the Boolean atoms, satisfying such a correspondence are found using off-the-shelf solvers. The set of solutions specifies the set of SMCMs that satisfy all the input CI/CDs and the background assumptions. Given that the results of the statistical tests may conflict with the background assumptions, there may be no solution, i.e.\ there is no SMCM that satisfies all the input CI/CDs and background assumptions. For that case \citet{hej2014} introduced the following optimization to resolve the conflict: 
\begin{align}
G^* &\in \arg \min_{G \in \mathcal{G}} \sum_{k \in \K \textrm{ s.t. } G \not \models k} w(k) \label{eq:optimization}
\end{align} 
In words, an output graph $G^*$ minimizes the weighted sum of input CI/CDs, which it does not satisfy given the encoded background assumptions. \citet{hej2014} adopted three weighting schemes for the weights $w(.)$: (1) ``constant weights" (CW) assigns a weight of 1 to each CI and CD constraint. (2) ``hard dependencies" (HW/NoI-m) assigns infinite weight to any observed CD, and a weight of 1 to any CI. (3) ``log weights" (LW) is a pseudo-Bayesian weighting scheme, where the weights depend on the log posterior probability of the CI/CDs being true (see their Sec.~4).  

To encode V-adjacency-faithfulness and V-adjacency-minimality, we need to encode in ASP what it is for an SMCM to have an inducing path and a virtual adjacency, and then replace the encoding of the Faithfulness assumption in \citet{hej2014} with its weaker versions.
Figure \ref{fig:asp-encoding} summarizes the ASP-encoding of V-adjacency-faithfulness and V-adjacency-minimality. We briefly explain the predicates:

\begin{itemize}
\item $edge(X,Z)$ and $conf(X,Z)$: $X\rightarrow Z$ and $X \leftrightarrow Z$, respectively, are in the SMCM.
\item $ancestors(Z,X,Y)$: $Z$ is an ancestor of $X$ or $Y$ in the SMCM.
 \item $h(X,Z,Y)$: There is a path between $X$ and $Z$ which is into $Z$, and if the path consists of two or more edges, every non-endpoint vertex on the path is a collider and every vertex on the path is an ancestor of either $X$ or $Y$.
\item $t(X,Z,Y)$: It differs from $h(X,Z,Y)$ only in that the path between $X$ and $Z$ is out of $Z$. Together, $t(.)$ and $h(.)$ are used to specify the possible inducing paths.
\item $vadj(X, Y)$: $X$ and $Y$ are virtually adjacent. 
\item $indep(X, Y, \C, w)$: $X$ and $Y$ are independent conditional on $C$, given as input fact, with weight $w$.
\end{itemize}

For V-adjacency-faithfulness, we encode that any CI statement $X \independent Y \given \C$ implies that $X$ and $Y$ are not virtually adjacent. For V-adjacency-minimality, we employ the minimization of the number of virtual-adjacencies. 
By encoding the weaker assumptions in the framework of \citet{hej2014}, we then have the following algorithms (Hyttinen et al.'s algorithm based on the `hard dependencies' weights is equivalent to one based on NoI-minimality):

\begin{itemize}
\item $\mathbf{VadjF}$: Virtual-adjacency-faithfulness + Markov
\item $\mathbf{VadjM}$: Virtual-adjacency-minimality + Markov
\end{itemize}

\newcommand{\asp}{\mbox{ :- }}                                                           \newcommand{\wasp}{ :\sim\, }
                      
\renewcommand{\land}{,}    
\newcommand{\naf}{\mbox{not }}

\begin{figure}
\vspace{-.3cm}
\begin{framed}
\hspace{-3mm}
\begin{minipage}{\columnwidth}
\small
{\bf Inference rules for virtual-adjacency:} \\
\vspace*{-3mm}
\hspace{-2mm}
$$
h(X,Z,Y) \asp edge(X,Z), ancestors(Z,X,Y).$$
\vspace*{-3mm}
\hspace{-2mm}
$$
h(X,Z,Y) \asp conf(X,Z), ancestors(Z,X,Y).$$
\vspace*{-3mm}
\hspace{-2mm}
$$
h(X,Z,Y) \asp h(X,U,Y), conf(Z,U),
ancestors(Z,X,Y).$$
\vspace*{-3mm}
\hspace{-2mm}
$$
t(X,Z,Y) \asp h(X,U,Y), edge(Z,U), ancestors(Z,X,Y).$$
\vspace*{-3mm}
\hspace{-2mm}
$$
vadj(X, Y) \asp  h(X,Y,Y).$$
\vspace*{-3mm}
\hspace{-2mm}
$$
vadj(X, Y) \asp  t(X,Y,Y).$$
\vspace*{-3mm}
\hspace{-2mm}
$$
vadj(X, Y) \asp  edge(Y,X).$$

{\bf Virtual-adjacency-faithfulness (violations):} \\
$\forall X\forall Y>X$, $\forall \C \subseteq \V\setminus \{X, Y\}$,
\vspace*{-3mm}
$$ 
\hspace{-2mm}\begin{array}{r@{}c@{}lr}
\asp & \naf vadj(X, Y), indep(X, Y, \C, w) &
\end{array}
$$

{\bf Virtual-adjacency-minimality (optimization of weak constraints)}: \\
$\forall X\forall Y>X$,
\vspace*{-3mm}
$$ 
\hspace{-2mm}\begin{array}{r@{}lr}
fail(X, Y, w=1) &\asp vadj(X, Y).\\
\end{array}
$$
$$
\begin{array}{l@{}l}
\wasp & \mathit{fail}(X,Y,w).\,[w]
\end{array}
$$

(Variables are in an arbitrary order so that $indep(X, Y, \C, w)$ and $dep(X, Y, \C, w)$ are considered only if $Y>X$, in order to avoid double counting.) 
\end{minipage}
\end{framed}
\caption{ASP Encoding of V-adjacency-faithfulness and V-adjacency-minimality \label{fig:asp-encoding}}
\end{figure}

\section{Simulations}
\label{sec: simulations}

We report two types of simulation, one using an independence oracle that specifies the true CI/CDs of the causal model, and one that uses the CI/CDs inferred from the sample data. 

For both simulations we followed the model generation process of \citet{hej2014} for causally insufficient models: We generated 100 random linear Gaussian models over 6 vertices with an average edge degree of 1 for directed edges. The edge coefficients were drawn uniformly from $[-0.8, -0.2]\cup [0.2, 0.8]$. The error covariance matrices (which also represent the confounding) were generated using the observational covariance matrix of a similarly constructed causally sufficient model (with its error covariances sampled from $N(0.5, 0.01)$). 



In the oracle setting, we randomly generated 100 linear Gaussian models with latent confounders over 6 variables and then input the independence oracles implied by these models. We observed that the algorithms based on V-adjacency-faithfulness, on V-adjacency-minimality, and on NoI-minimality (which is equivalent to using `hard dependencies' weighting) all returned the exact same results as the algorithm based on Faithfulness did, which is consistent with the theoretical results in Section \ref{sec: weakeningSMCM} and confirms the correctness of our encoding.



In the finite sample case we generated five data sets with 500 samples from each of the 100 models. We used correlational t-tests and tried 10 threshold values for rejecting the null hypothesis ($0.0001$, $0.0005$, $0.001$, $0.005$, $0.01$, $0.05$,  $0.1$, $0.15$, $0.2$,  $0.25$). The test results formed the input for the algorithms. We also used the log-weighting scheme and tried 10 values for the free parameter of the Bayesian test ($0.05, 0.09, 0.1, 0.15, 0.2, 0.3, 0.4, 0.5, 0.6, 0.7, 0.9$).

\begin{figure}[h]
\centerline{
{\includegraphics[width=0.55\textwidth]{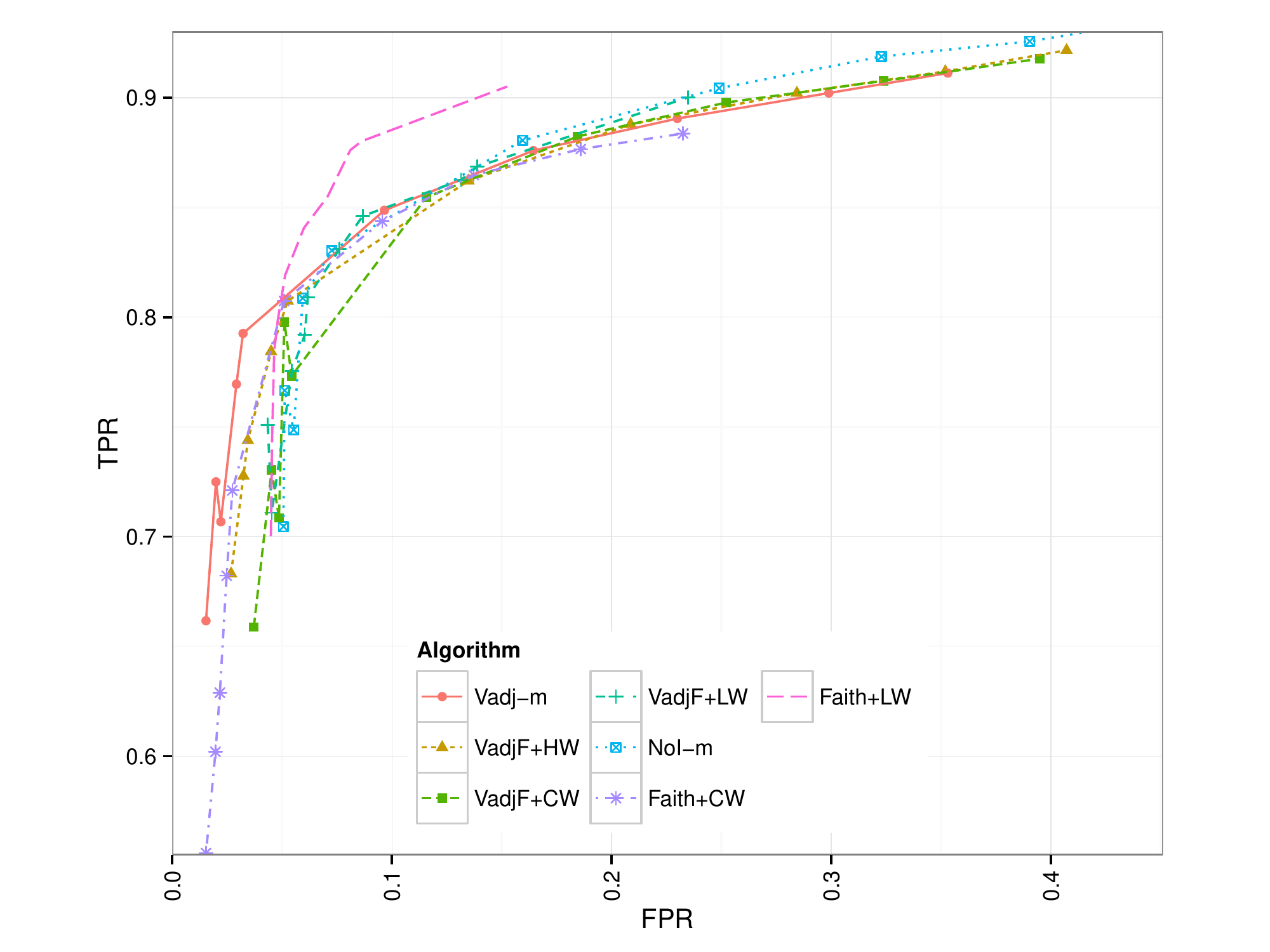}}
}
\caption{ROC of d-connections}
\label{fig: roc}
\end{figure}

For each algorithm we output all possible solutions and compared the d-connections common to all the output graphs against those of the true data generating graph. In all the $100$(models)$*5$(datasets)$*10$(parameters) $=5,000$ runs, Faithfulness was satisfied in only $367/5000$ of the cases while V-adjacency-faithfulness was satisfied in $2065/5000$ of the cases. This shows that V-adjacency-faithfulness is indeed significantly weaker than Faithfulness and greatly reduces the number of conflicts. 
By definition, V-adjacency-minimality can always be satisfied. Figure \ref{fig: roc} plots the ROC curves for the inferred d-connections. Under ``constant weighting'' (CW) ``hard dependencies weighting'' (HW), using V-adjacency-faithfulness achieves comparable accuracy to using faithfulness, with some trade-offs between false-positive rates and true-positive rates. Under ``log weighting'' (LW), however, using Faithfulness seems slightly more accurate than using V-adjacency-faithfulness, though using V-adjacency-minimality seems to generally yield the lowest false-positive rates. How to adapt the ``log weighting'' to fit V-adjacency-faithfulness better is an interesting question for future work. 

\begin{figure}[h]
\centerline{
{\includegraphics[width=0.5\textwidth]{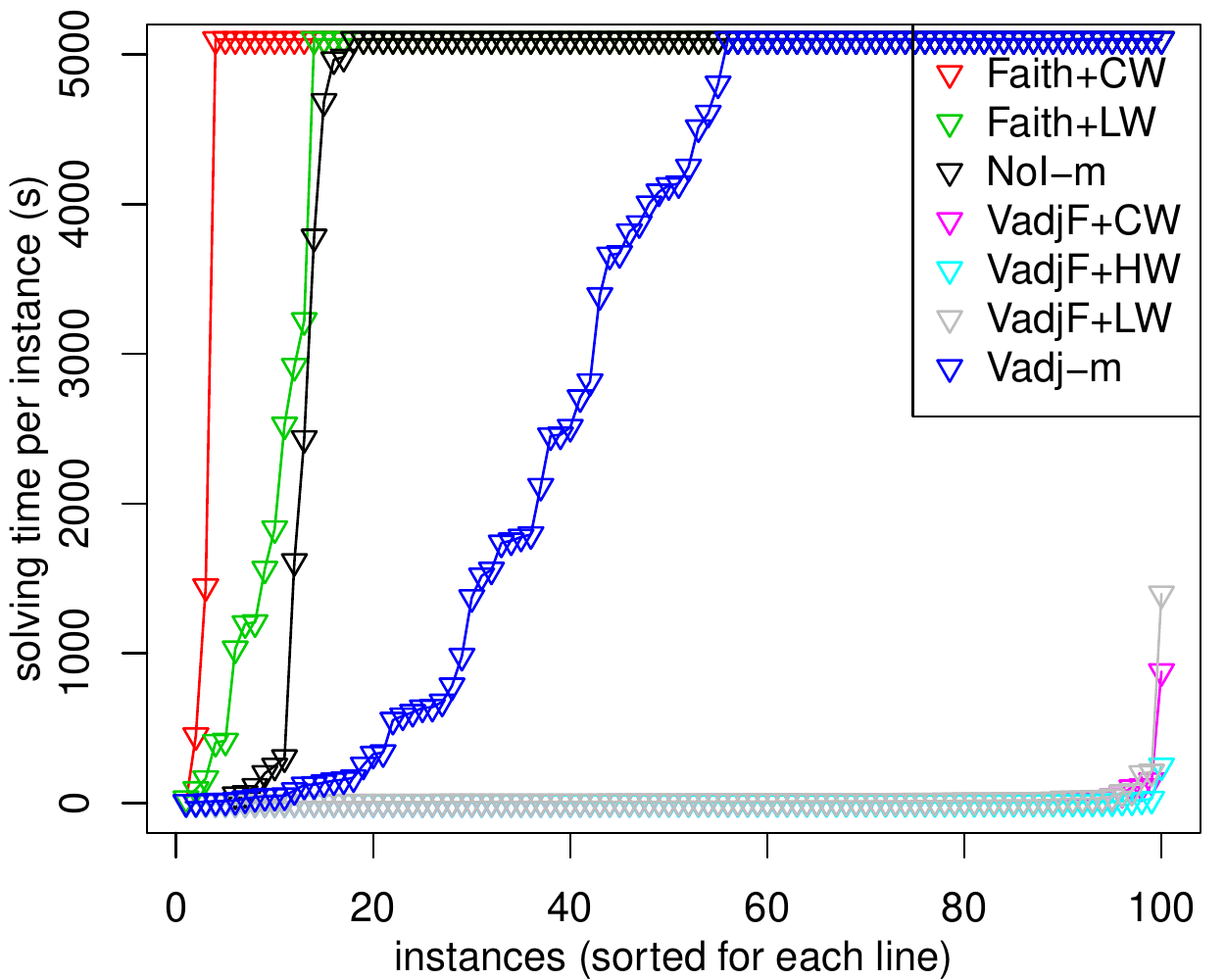}}
}
\caption{Sorted Solving Times for 8 Variables (time-out at 5,000s)}
\label{fig: SLVtm}
\end{figure}

Finally, to explore the efficiency gain of the weakened faithfulness assumptions, we generated 100 random linear Gaussian models with latent confounders over 8 variables and generated one data set with 500 samples from each model. For each algorithm, we only required that one graph be found. Figure \ref{fig: SLVtm} shows the sorted solving times for the different background assumptions (with maximum time budget of 5,000s).
As in the causally sufficient case, we see a significant improvement in solving times when using the weakened faithfulness assumptions.

\section{Conclusion}
\label{sec: conclusion}
We have shown how to extend the results on weakening Faithfulness in the context of learning causal DAGs to the more realistic context of learning SMCMs that allow for the representation of unmeasured confounding. We identified generalizations of some proposals of weakening Faithfulness in the literature and showed that they continue to be what we call conservative weakenings. Moreover, we implemented ASP-based algorithms for learning SMCMs based on these weaker assumptions. The simulation results suggest that some of these weaker assumptions, especially V-adjacency-faithfulness, help to save solving time in ASP-based algorithms to a significant extent.  


In this connection, a direction of future work is to explore how the apparent advantage of using weaker assumptions may be realized on top of other ASP-based causal discovery methods, such as ETIO in \citep{Borboudakis2016} and ACI in \citep{magliacane2016ancestral}. 

\begin{figure}[h]
\centerline{
{\includegraphics[width=0.5\textwidth]{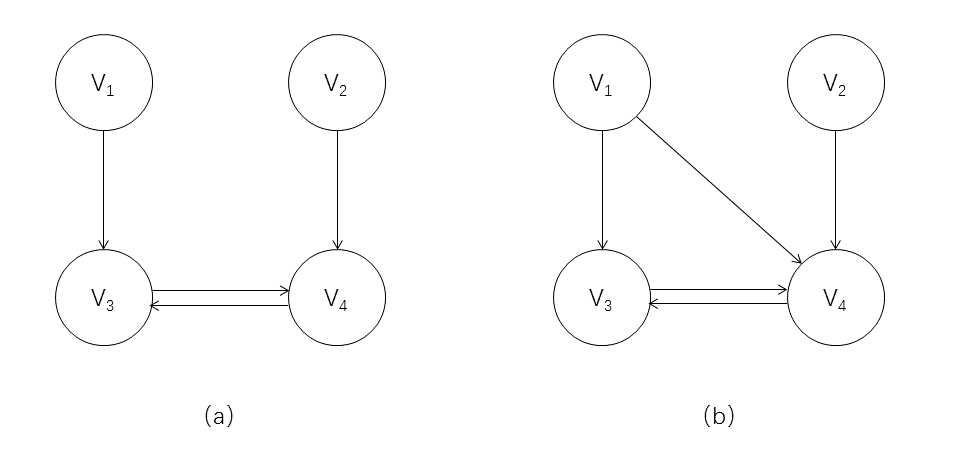}}
}
\caption{Illustration of a complication in cyclic models}
\label{fig: cyclic}
\end{figure}

One great appeal of the ASP-based approach is that the background assumptions that determine the search space can be flexibly adjusted to include causal models with both latent confounding and causal feedback. We close with an illustration of a (further) complication that arises in cyclic causal models. Suppose the true causal structure is the cyclic one in Figure \ref{fig: cyclic}(a), which entails that $V_1\dsep V_2$ and $V_1\dsep V_2\given \{V_3, V_4\}$. Suppose the true distribution is Markov and Faithful to this structure and hence features exactly two nontrivial conditional independencies. Then the distribution is not Faithful to the structure in Figure \ref{fig: cyclic}(b) (for that structure does not entail $V_1\dsep V_2\given \{V_3, V_4\}$), but it is still V-adjacency-faithful (for $V_1$ and $V_2$ are not virtually adjacent).        

This means that even V-adjacency-faithfulness is not a conservative weakening of Faithfulness when causal feedback is allowed. Whether it can be strengthened into a useful conservative weakening for the purpose of learning cyclic models is worth further investigation. 

\section*{Acknowledgments}
JZ was supported by GRF LU13602818 from the RGC of Hong Kong. FE was supported by NSF grant 1564330.


\begin{appendices}

\appendix

Below we prove the theorem stated in the paper. The result is restated separately as three theorems in Section C below, one for each of the three weakenings of the Faithfulness assumption we considered. The proof makes use of maximal ancestral graphs (MAGs), for which useful characterizations of Markov equivalence are available. Below we proceed as follows. In Section A, we introduce MAGs and known facts that are relevant to our arguments. In Section B, we describe a connection between SMCMs and MAGs that we will exploit. Finally, in Section C, we prove the theorems.


\section{Maximal Ancestral Graphs}

Like an SMCM, a MAG is a graphical object designed to represent a causal structure in the presence of latent variables. An {\it Ancestral graph} is a {\it simple} mixed graph (i.e., at most one edge can appear between any two vertices), in which for any two vertices $X$ and $Y$, if $X$ is an ancestor of $Y$, then there is no (directed or bi-directed) edge between $X$ and $Y$ that is into $X$. A {\it Maximal Ancestral Graph} is an ancestral graph in which for every pair of non-adjacent vertices, there exists some set of other vertices that m-separates them. A DAG is a special MAG which does not contain bi-directed edges.

Two DAGs are Markov equivalent if and only if they share the same adjacencies and unshielded colliders. Although Markov equivalent MAGs also share the same adjacencies and unshielded colliders, these commonalities are no longer sufficient to characterize Markov equivalence between MAGs. Two Markov equivalent MAGs may also share some shielded colliders. The definitions related to this fact are given below. 
 
 \theoremstyle{definition}
 \begin{definition}
\citep{Richardson02ancestralgraph} A path $\pi = (V_0, V_1,...,V_m = X, Z, Y) (m \geq 1) $ is a {\it discriminating path for} $ \langle X,Z,Y \rangle $ in a MAG, if $V_0$ and $Y$ are not adjacent and every vertex $V_i, 1\leq i\leq m$, is a collider on $\pi$ and a parent of $Y$.
\end{definition}



 \begin{definition}
An {\it inducing path} $\pi$ between $ X$ and $Y$ is a path on which every non-endpoint vertex is a collider and an ancestor of either $X$ or $Y$.
 \end{definition}

 \begin{definition}
\citep{ali2009}: Call $\langle X, Z, Y\rangle$ a triple if $X, Z$ are adjacent and $Y, Z$ are adjacent. The order of such a triple in a MAG is defined recursively as follows: 

{\it Order}  $0$. A triple $ \langle X,Z,Y \rangle $ has order 0 if $X$ and $Y$ are not adjacent. 

{\it Order}  $i$. A triple $ \langle X,Z,Y \rangle $ has order $i$, if it does not have any order less than $i$, and there is a discriminating path 
$ \langle  V_0, V_1,...,V_m = X, Z, Y\rangle $ or $ \langle V_0, V_1,...,V_m = Y, Z, X\rangle $, on which every collider triple centered at $V_j$ ($1\leq j\leq m$) has order at most $i-1$ (and at least one of them has order $i-1$).

 \end{definition}

A discriminating path for $\langle X,Z,Y \rangle$ is said to {\it have order $i$} if except for $\langle X,Z,Y \rangle$, every collider triple on the path has order less than $i$ and at least one of them has order $i-1$. Note that some triples in a graph may not have an order. Note also that the order (if any) of a shielded triple is the minimum of all discriminating paths with order for that triple \citep{ali2009}. Colliders with order $ \geq 1$ are those shielded colliders that are present in all Markov equivalent MAGs. \\
\\
{\bf Proposition 1 } \citep{ali2009} Two MAGs are Markov equivalent if and only if 
they have the same adjacencies and the same colliders with order.

\section{SMCMs and MAGs}


SMCMs and MAGs are both generalizations of DAGs. Like in DAGs, directed cycles are not allowed in SMCMs or MAGs. But unlike DAGs, they can contain bi-directed edges ($\bia$) in addition to the directed edges ($\ra$). However, almost directed cycles --- where one endpoint of a bi-directed edge is an ancestor of the other endpoint --- are allowed in SMCMs but not in MAGs. For this reason at most one edge is allowed between any two variables in a MAG, while in SMCMs up to two edges (one directed and one bi-directed) are allowed. Besides, the interpretation of an edge in an SMCM is different from that in a MAG. A directed edge $X \ra Y$ in an SMCM means that $X$ is a direct cause of $Y$ relative to $\V$. A bi-directed edge $X \bia Y$ means that $X$ and $Y$ are confounded by a latent variable. In contrast, in a MAG a directed edge $X \ra Y$ represents that $X$ is a causal ancestor of $Y$, and a bi-directed edge $X \bia Y$ means that $X$ is not an ancestor of $Y$ and $Y$ is not an ancestor of $X$ (which then imply that they are confounded by a latent variable). If $X$ is a causal ancestor of $Y$ (that is not mediated by any other observed variable), and they also have a common cause which is latent, then only a directed edge $X \ra Y$ is present in the corresponding MAG.

For a causally insufficient system $\V$, we assume that there is a causal DAG $G$ over $\V \cup \mathbf{L}$ (where $\mathbf{L}$ is a set of latent variables) that satisfies the causal Markov assumption. The set of m-separations (d-separations) entailed by $G$ is called the independence model associated with $G$ and denoted as $\mathbf{J}(G)$. The marginal independence model of $\mathbf{J}(G)$ over $\V$ after leaving out the set of latent variables $\mathbf{L}$ is the subset of m-separations (d-separations) entailed by $G$ which do not involve any variables in $\mathbf{L}$. We denote it as $\mathbf{J}(G)|_\mathbf{L}$. Given $G$ over $\V \cup \mathbf{L}$, there is a unique MAG $M$ and a unique SMCM $S$ over $\V$, which represent some causal relations among $\V$ in $G$ and the marginal independence model of $G$ over $\V$. 

A DAG over $\V$ with latent variables $\mathbf{L}$ can be projected into such an SMCM in the following way \citep{Verma1993, TianP02}:\\
\\
\noindent{\it Input}: a DAG $G$ over $\V \cup \mathbf{L}$ 

\noindent{\it Output}: an SMCM $S_G$ over $\V$

\begin{enumerate}
\item Add each variable in $\V$ as a node of $S_G$.
\item For each pair of variables $X, Y \in \V$, if there is an edge between them in $G$, add the edge to $S_G$.
\item For each pair of variables $X, Y \in \V$, if there is a directed path from $X$ to $Y$ in $G$ such that every mid node on the path is in $\mathbf{L}$, add edge $X \ra Y$ to $S_G$, if it does not exist yet.
\item For each pair of variables $X, Y \in \V$, if there exist a directed path from a variable $L_i \in \mathbf{L}$ to $X$ and a directed path from $L_i$ to $Y$ in $G$ such that every mid node on the paths is in $\mathbf{L}$, add edge $X \bia Y$ to $S_G$, if it does not exist already.
\end{enumerate}


The conversion of a DAG with latent variables to a MAG is given in \citet{Richardson02ancestralgraph}.

Given a set of variables $\V$,  For every SMCM over $\V$, there is also a unique MAG that corresponds to it, such that they entail the same set of conditional independence statements (and the causal relations represented by the MAG are compatible with those represented by the SMCM). The following is a procedure to transform an SMCM into its corresponding MAG. It is adapted from the algorithm presented in \citet{DBLP:journals/jmlr/Zhang08}, which is used to project DAGs with latent variables to MAGs. \\
\\
{\bf Conversion from an SMCM to a MAG}

\begin{steps}
   \item For each pair of variables $X$ and $Y$, $X$ and $Y$ are adjacent in the output MAG $M$, if there is an inducing path between them in the input SMCM $S$.
   
  \item For each pair of adjacent variables $X$ and $Y$ in the output MAG $M$,
  \begin{enumerate}[label=(\roman*)]
  \item If $X$ is an ancestor of $Y$ in $S$, orient the edge as $X \ra Y$.
  \item If $Y$ is an ancestor of $X$ in $S$, orient the edge as $Y \ra X$. 
  \item else, orient the edge as $X \bia Y$.
  \end{enumerate}
 
\end{steps}

\begin{figure}
\centerline{
{\includegraphics[width=0.5\textwidth]{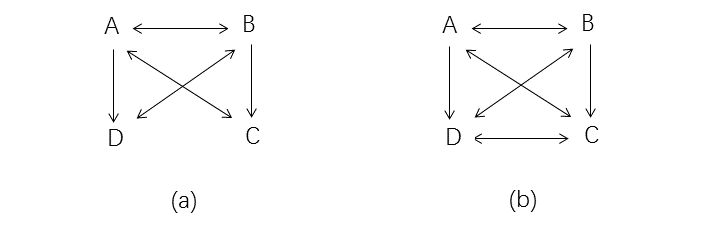}}
}
\caption{An SMCM (left) and its corresponding MAG (right) in which $C$ and $D$ are not m-separated by any subset of $\{A, B\}$.}
\label{fig: SMCMandMAG}
\end{figure}

It is worth noting that in MAGs, two variables are adjacent if and only if they are not m-separated by any set of other variables, This is not true for SMCMs,  as shown in Figure~\ref{fig: SMCMandMAG}. However, in SMCMs it holds that there is an inducing path between two variables if and only if the two variables are not m-separated by any set of other variables. For example, in Figure~\ref{fig: SMCMandMAG}(a), there is an inducing path ($C$, $A$, $B$, $D$) between $C$ and $D$.
We write the property as a proposition for later reference.  \\


\noindent {\bf Proposition 2 } If $S$ is an SMCM over $\V$, $M$ is the corresponding MAG of $S$ over $\V$, and $X, Y \in \V$, then the following statements are equivalent:

\begin{enumerate}[label=\arabic*)]
\item 
There is an inducing path between $X$ and $Y$ in $S$.

\item $X$ and $Y$ are m-connected given any subset of $\V \setminus \{X, Y\}$.

\item $X$ and $Y$ are adjacent in $M$.

\end{enumerate}

{\it Proof}.  A proof of the equivalence of 2) and 3) is given in the proof of Theorem 4.2 in \citet{Richardson02ancestralgraph}. The equivalence of 1) and 3) can be proved in the same way as the equivalence of (ii) and (iii) of Theorem 4.2 in \citet{Richardson02ancestralgraph}.

\section{Conservative Weakenings of Faithfulness}
In this section, we prove the theorem in the paper, via generalizations of the weakenings of Faithfulness to MAGs and establishing related results. In MAGs, it holds that if two vertices $X$ and $Y$ are adjacent, then there is no subset of $\V \setminus \{X, Y\}$ that m-separates them. Hence, the formulation of the Adjacency-faithfulness established for DAGs can be directly applied to MAGs, except that the definition of adjacency allows for bi-directed edges. Likewise, the formulation of 
Number-of-Edges(NOE)-minimality carries over to MAGs except that bi-directed edges are allowed in MAGs. 
The Number-of-Independences(NoI)-minimality assumption, which is given in terms of conditional independence statements, can be directly extended to MAGs. In the following, we prove that for MAGs, all of the generalized assumptions remain conservative weakenings of Faithfulness. Based on these results, we prove the corresponding theorems for SMCMs. \\
\\
\noindent {\bf Lemma 1 } Given the causal Markov assumption, the Adjacency-faithfulness assumption is a conservative weakening of the Faithfulness assumption in the case of MAGs, in the following sense:
\begin{enumerate}
    \item[(a)] the Adjacency-faithfulness assumption is entailed by, but does not entail, the Faithfulness assumption. 
    \item[(b)] For every joint probability distribution $P$ over $\V$, if there exists a MAG that satisfies both Markov and Faithfulness assumptions with $P$, then for every MAG $G$ that satisfies the Markov assumption with $P$, $G$ satisfies Faithfulness if and only if $G$ satisfies the Adjacency-faithfulness assumption with $P$.   
\end{enumerate}


\begin{proof}

(a) 
Let $G_F$ be a MAG over $\V$ to which $P$ is both Markov and faithful. If two variables $X$ and $Y$ are adjacent in $G_F$, they are not m-separated given any set $\C \subseteq \V \setminus \{X, Y\}$. Then, since $P$ is faithful to $G_F$, $X$ and $Y$ are not independent given any set $\C \subseteq \V \setminus \{X, Y\}$ in $P$. Thus, $G_F$ satisfies Adjacency-faithfulness. Therefore, the Adjacency-faithfulness assumption is entailed by the Faithfulness assumption.


We show that Adjacency-faithfulness does not entail Faithfulness with an example. Suppose that $\mathbf{V}=\{V_1,V_2,V_3\}$ and the conditional independence relations satisfied by the distribution are $V_1\independent V_3$ and $V_1\independent V_3 \given V_2$ (see Zhang, 2013 for an example of such a distribution.) Then given Markov, the structure $V_1\ra V_2\lra V_3$ satisfies Adjacency-faithfulness but not Faithfulness with the distribution.


(b) Suppose there exists a MAG $G_F$ that is both Markov and faithful to $P$. The ``only if" direction has already been proved in (a), so we just need to prove that, in this case, a MAG satisfies the Faithfulness assumption with $P$ if it satisfies Adjacency-faithfulness assumption with $P$. 
In other words, we just need to prove that in this case, every MAG $G_A$ that satisfies Markov Adjacency-faithfulness with $P$ is Markov equivalent to $G_F$. 

By Proposition 1, the MAGs that are Markov equivalent have the same adjacencies and colliders with order.
It is easy to see that $G_A$ shares the same adjacencies with $G_F$. The proof given for DAGs \citep{ramsey06} is directly applicable to MAGs. So we only need to prove that $G_A$ and $G_F$ have the same colliders with order.
For this purpose, we first prove the following claim.\\

\noindent {\bf Claim}: Let $\pi = (V_0, V_1,...,V_m = X, Z, Y) (m \geq 1)$ be a discriminating path for $Z$ in a MAG $G_F$ which satisfies the Markov and the Faithfulness assumption with $P$. If the corresponding path $\pi^*$ forms a discriminating path in a MAG $G_A$ that satisfies Markov with $P$, then $Z$ is a collider on $\pi$ in $G_F$ if and only if $Z$ is a collider on $\pi^*$ in $G_A$.

\begin{proof}
 
As every vertex $V_i,1\leq i\leq m$ is a collider on path $(V_0,V_1,...,V_m=X, Z, Y)$ and a parent of $Y$, any set that m-separates $V_0$ and $Y$ must include all of them. 

If $(X, Z, Y)$ is a collider in $G_F$, the path $(V_0, V_1,...,V_m = X, Z, Y)$ m-connects given any superset of $\mathbf{S}=\{V_1,...,V_m\}\cup\{ Z\}$. As $G_F$ satisfies the Faithfulness assumption with $P$, $V_0$ and $Y$ are dependent given any superset of $\mathbf{S}$ in $P$. If $(X, Z, Y)$ is not a collider in $G_A$, 
there must be some superset of $\mathbf{S}$ that m-separates $V_0$ and $Y$.  But then $G_A$ would violate the Markov assumption, which is a contradiction. Hence, $(X, Z, Y)$ is also a collider in $G_A$.

If $(X, Z, Y)$ is not a collider in $G_F$, then the path $(V_0,V_1,...,V_m=X, Z, Y)$ m-connects given any superset of $\{V_1,...,V_m\}$ which does not include $Z$. By Faithfulness, $V_0$ and $Y$ are dependent given any superset of $\{V_1,...,V_m\}$ which does not include $Z$ in $P$. If $(X, Z, Y)$ is a collider in $G_A$, any set that m-separates $V_0$ and $Y$ must be a superset of $\{V_1,...,V_m\}$ which does not include $Z$. But then $G_A$ would violate the Markov assumption, which is a contradiction. 
Thus, $(X, Z, Y)$ is not a collider in $G_A$.

\end{proof}

Now we can prove that a triple $(X, Z, Y)$ is a collider with order r in $G_F$ if and only if $(X, Z, Y)$ is a collider with order r in $G_A$.

Let $r$ be the order of $(X, Z, Y)$.

If $r=0$, $(X, Z, Y)$ is an unshielded collider. The proof  given for DAGs in \citet{ramsey06} is directly applicable here. The only difference is that in MAGs, colliders and non-colliders admit more edge configurations than they do in DAGs. 

When $r>0$, assume the result holds for all $s < r$. 

If $(X, Z, Y)$ is a triple with order $r$ in $G_F (G_A)$, by the definition of ordered triple, there exists a discriminating path $\pi = (V_0,V_1,...,V_m=X, Z, Y)$ (or $\pi = (V_0,V_1,...,V_m=Y, Z, X)$) in $G_F (G_A)$ such that, except $(X, Z, Y)$, every triple $V_i$  $( 1\leq i\leq m) $ on $\pi$ is a collider and has order less than $r$. And since $G_F$ and $G_A$ have the same adjacencies, the sequence of vertices forming the discriminating path in $G_F (G_A)$, also forms a path in $G_A (G_F)$. Let $\pi^*$ be the corresponding path in $G_A (G_F)$. By the inductive hypothesis, in $G_A (G_F)$, each collider $V_i$  $( 1\leq i\leq m) $ is also a collider with the same order as in $G_F (G_A)$ on the corresponding path $\pi^*$. 
We claim that the corresponding path $\pi^*$ is also a discriminating path in $G_A (G_F)$ for $(X, Z, Y)$. Since we have $V_0 ? \ra V_1 \lra ... \lra V_m \la ? Z $ in $G_A (G_F)$, it suffices to show that $V_j \ra Y ( 1\leq j\leq m) $ in $G_A (G_F)$. 

Triple $(V_0 , V_1 , Y)$ is a noncollider with order $0$ in $G_F (G_A)$, because $V_0$ and $Y$ are not adjacent. Hence, $(V_0 , V_1 , Y)$ is also a noncollider with order $0$ in $G_A (G_F)$. Further, as $V_0 ? \ra V_1 $, by the definition of MAGs, $V_1 \ra Y $ in $G_A (G_F)$. Arguing inductively, assume 
$V_i \ra Y ( 1 < i < m) $ in $G_A (G_F)$, so that $(V_0,V_1,...,V_m,Y)$ forms a discriminating path with order at most $r$ for  $(V_{m-1} , V_m , Y)$ in both $G_A$ and $G_F$. As a consequence, as $(V_{m-1} , V_m , Y)$ is a noncollider on $(V_0,V_1,...,V_m,Y)$ in  $G_F (G_A)$, $(V_{m-1} , V_m , Y)$ is a noncollider on $(V_0,V_1,...,V_m,Y)$ in  $G_A (G_F)$, based on the claim established above. Since $V_{m-1} \ra V_m $, by the definition of MAGs, $V_m \ra Y $ in $G_A (G_F)$. Hence, $\pi^*$ also forms a discriminating path in $G_A (G_F)$ for $(X, Z, Y)$. Again, based on Lemma 4, $(X, Z, Y)$ is a collider in $G_F$ if and only if $(X, Z, Y)$ is a collider in $G_A$.

By the definition of an ordered triple, $(X, Z, Y)$ has order at most $r$ in $G_A$. However, if $(X, Z, Y)$ has order less than $r$ in $G_A$, by the inductive hypothesis, $(X, Z, Y)$ will have order less than $r$ in $G_F$, which is a contradiction. Thus, $(X, Z, Y)$ has order $r$ in both graphs. 


To summarize, $G_A$ and $G_F$ are Markov equivalent since they have the same adjacencies and colliders with order.
\end{proof}

\noindent {\bf Theorem 1 }Given the causal Markov assumption, the V-adjacency-faithfulness assumption is a conservative weakening of the Faithfulness assumption in the case of SMCMs, in the following sense: 

\begin{enumerate}
    \item[(a)] V-adjacency-faithfulness is entailed by, but does not entail, Faithfulness. 
    \item[(b)] For every joint probability distribution $P$ over $\V$, if there exists an SMCM that satisfies both Markov and Faithfulness assumptions with $P$, then for every SMCM $G$ that satisfies the Markov assumption with $P$, $G$ satisfies Faithfulness if and only if $G$ satisfies the V-adjacency-faithfulness with $P$.   
\end{enumerate}


\begin{proof}

(a) Let $S_F$ be an SMCM which satisfies Markov and Faithfulness with $P$. By Proposition 2, if two variables $X$ and $Y$ are virtually adjacent in $S_F$, they are m-connected given any subset of $\V \setminus \{X, Y\}$. Then by Faithfulness, $X$ and $Y$ are dependent conditional on any subset of $\V \setminus \{X, Y\}$. So $S_F$ satisfies V-adjacency-faithfulness. Hence, the V-adjacency-faithfulness assumption is entailed by the Faithfulness assumption.

However, V-adjacency-faithfulness does not entail Faithfulness. It can be illustrated with the same example used in the proof of Lemma 1, since syntactically MAGs are special cases of SMCMs.  

(b) Now we prove the ``if" direction, since the ``only if" direction has already been proved in (a). Let $S_I$ be an SMCM which satisfies Markov and V-adjacency-faithfulness with $P$ and $M_I$ be the unique MAG corresponding to $S_I$. 

If $X$ and $Y$ are adjacent in $M_I$, there is an inducing path between $X$ and $Y$ in $S_I$, based on Proposition 2. Then, $X$ and $Y$ are not independent given any subset of $\V \setminus \{X, Y\}$, since $S_I$ satisfies V-adjacency-faithfulness. Thus, $M_I$ satisfies Adjacency-faithfulness. And, if $P$ is faithful to some SMCM, it is faithful to the corresponding MAG of that SMCM, which means that there exists a MAG that satisfies both Markov and Faithfulness with $P$. Further, by Lemma 1, $M_I$ satisfies the Faithfulness assumption. It follows that $S_I$ satisfies the Faithfulness assumption, since $M_I$ and $S_I$ entail exactly the same CI statements.

\end{proof}

\noindent {\bf Lemma 2 }Given the causal Markov assumption, the NOE-minimality assumption is a conservative weakening of the Faithfulness assumption in the case of MAGs, in the following sense:
\begin{enumerate}
    \item[(a)] the NOE-minimality assumption is entailed by, but does not entail, the Faithfulness assumption. 
    \item[(b)] For every joint probability distribution $P$ over $\V$, if there exists a MAG that satisfies both Markov and Faithfulness assumptions with $P$, then for every MAG $G$ that satisfies the Markov assumption with $P$, $G$ satisfies Faithfulness if and only if $G$ satisfies the NOE-minimality assumption with $P$.   
\end{enumerate}


\begin{proof}
(a) Let $G_F$ be a MAG to which $P$ is both Markov and faithful. Then removing any edge from $G_F$ will either violate the maximality or introduce an independence which is not satisfied by $P$, which constitutes a violation of the Markov assumption. Hence, there is no MAG with a smaller number of edges than $G_F$, which satisfies Markov. So $G_F$ satisfies NOE-minimality. Thus, NOE-minimality asusmption is entailed by the Faithfulness assumption. 

\citet{ForsterEtal2017} showed that NOE-minimality is weaker than Faithfulness for DAGs. Thus, in the case of MAGs, NOE-minimality does not entail Faithfulness, since DAGs are special cases of MAGs.

(b) Suppose there exists a MAG $G_F$ that is both Markov and faithful to $P$, and suppose $G_{NoE}$ is a MAG that satisfies Markov and NOE-minimality with the distribution $P$. As we did previously, to prove the ``if" direction, we only need to prove that $G_F$ and $G_{NoE}$ have the same adjacencies and colliders with order.

$G_F$ and $G_{NoE}$ have the same number of edges since $G_F$ also satisfies NOE-minimality. Now we prove that $G_F$ and $G_{NoE}$ not only have the same number of edges but also the same adjacencies: For a contradiction, if we assume that $G_{NoE}$ has one different edge than $G_F$ does, then one edge that is present in $G_F$ is removed in $G_{NoE}$, since they share the same number of edges.  
As already mentioned, removing any edge that is present in $G_F$ will result in a  violation of either maximality or Markov. Thus, $G_F$ and $G_{NoE}$ not only have the same number of edges but also the same adjacencies. 

Next, we prove that $G_F$ and $G_{NoE}$ have the same unshielded colliders.
Since $G_F$ and $G_{NoE}$ have the same adjacencies, a triple is unshielded in $G_F$ if and only if it is unshielded in $G_{NoE}$. If an unshielded triple $(X, Z, Y)$ is an unshielded collider in $G_F$, then $X$ and $Y$ are dependent given any set that includes $Z$ in $P$, because the distribution $P$ is faithful to $G_F$. Then, as $P$ and $G_{NoE}$ satisfy the Markov assumption, $(X, Z, Y)$ is also an unshielded collider in $G_{NoE}$. Similarly, if $(X, Z, Y)$ is an unshielded non-collider in $G_F$, then it is an unshielded non-collider in $G_{NoE}$. 

The fact that $G_F$ and $G_{NoE}$ have the same colliders with order 
can be proved in the same way as we did in the proof of Lemma 1, since we have already proved that $G_F$ and $G_{NoE}$ have the same adjacencies and unshielded colliders.

\end{proof}


\noindent {\bf Theorem 2 }Given the causal Markov assumption, the V-adjacency-minimality assumption is a conservative weakening of the Faithfulness assumption in the case of SMCMs, in the following sense:
\begin{enumerate}
    \item[(a)] V-adjacency-minimality is entailed by, but does not entail, Faithfulness. 
    \item[(b)] For every joint probability distribution $P$ over $\V$, if there exists an SMCM that satisfies both Markov and Faithfulness assumptions with $P$, then for every SMCM $G$ that satisfies the Markov assumption with $P$, $G$ satisfies Faithfulness if and only if $G$ satisfies V-adjacency-minimality with $P$.   
\end{enumerate}


\begin{proof}

(a) 
Let $S_{AF}$ be an SMCM which satisfies Markov and V-adjacency-faithfulness with $P$. Then if two variables $X$ and $Y$ are virtually adjacent in $S_{AF}$, they are dependent given any subset of $\V \setminus \{X, Y\}$ in the distribution $P$. Then, by Proposition 2, taking away the virtual adjacency between $X$ and $Y$ will introduce a new conditional independence, which is not satisfied by the distribution $P$ and would thus result in a  violation of the Markov assumption. So $S_{AF}$ satisfies V-adjacency-minimality, which means that V-adjacency-minimality is entailed by V-adjacency-faithfulness. Further, V-adjacency-minimality is entailed by Faithfulness since V-adjacency-faithfulness is entailed by Faithfulness. 

However, V-adjacency-minimality does not entail the Faithfulness assumption because V-adjacency-faithfulness does not entail Faithfulness.

(b) Now we only need to prove the ``if" direction of (b) since the ``only if" direction has already been proved above. Let $S_{VADJ}$ be an SMCM which satisfies Markov and V-adjacency-minimality and $M_{VADJ}$ be the unique MAG corresponding to $S_{VADJ}$. 
 
By Proposition 2, there is an inducing path in $S_{VADJ}$ if and only if there is an edge in $M_{VADJ}$. If $M_{VADJ}$ does not satisfy NOE-minimality, there must be some MAG $M$, which has fewer edges than $M_{VADJ}$ and still satisfies Markov. The SMCMs corresponding to $M$ then have fewer virtual-adjacencies than $S_{VADJ}$ and also still satisfy Markov. This violates our initial assumption that $S_{VADJ}$ satisfies V-adjacency-minimality. Thus, $M_{VADJ}$ satisfies NOE-minimality. And, when $P$ is faithful to some SMCM, it is faithful to the corresponding MAG of this SMCM, which means that there exists a MAG that satisfies both Markov and Faithfulness with $P$. Further, by Lemma 2, $M_{VADJ}$ satisfies faithfulness, since it satisfies NOE-minimality. Hence $S_{VADJ}$ satisfies faithfulness, since it entails the same exact CIs with $M_{VADJ}$.
\end{proof}

\noindent {\bf Lemma 3 } Given the causal Markov assumption, the NOI-minimality assumption is a conservative weakening of the Faithfulness assumption in the case of MAGs, in the following sense:
\begin{enumerate}
    \item[(a)] the NOI-minimality asusmption is entailed by, but does not entail, the Faithfulness assumption. 
    \item[(b)] For every joint probability distribution $P$ over $\V$, if there exists a MAG that satisfies both Markov and Faithfulness assumptions with $P$, then for every MAG $G$ that satisfies the Markov assumption with $P$, $G$ satisfies Faithfulness if and only if $G$ satisfies the NOI-minimality assumption with $P$.   
\end{enumerate}


\begin{proof}
(a) The proof that was given for DAGs \citep{Zhalama2017SATBasedCD} is directly applicable to MAGs.

(b) Let $G_F$ be one of the graphs to which $P$ is both Markov and faithful and $G_{NoI}$ be a graph that satisfies Markov and NOI-minimality with $P$. Since $G_F$ also satisfies NOI-minimality, $G_F$ and $G_{NoI}$ entail the same number of conditional independence statements (CIs). Now we prove that they entail exactly the same CIs. For a contradiction, let's assume that $G_{NoI}$ entails one CI that is not entailed by $G_F$. Because $G_F$ satisfies faithfulness and Markov with $P$, the CIs entailed by $G_F$ are exactly the ones satisfied by $P$. But since $G_{NoI}$ entails one CI that is not satisfied by $P$, it then must violate Markov, which is a contradiction.  
Therefore, $G_F$ and $G_{NoI}$ entail the exact same CIs, which means that $G_{NoI}$ also satisfies Markov and faithfulness with $P$. 
\end{proof}

\noindent {\bf Theorem 3 } Given the causal Markov assumption, the NOI-minimality assumption is a conservative weakening of faithfulness in the case of SMCMs, in the following sense:
\begin{enumerate}
    \item[(a)] NOI-minimality is entailed by, but does not entail, Faithfulness. 
    \item[(b)] For every joint probability distribution $P$ over $\V$, if there exists an SMCM that satisfies both Markov and Faithfulness assumptions with $P$, then for every SMCM $G$ that satisfies the Markov assumption with $P$, $G$ satisfies Faithfulness if and only if $G$ satisfies NOI-minimality with $P$.   
\end{enumerate}


\begin{proof}
The proof of Lemma 3 can be directly extended to SMCMs.
\end{proof}


\end{appendices}

\bibliographystyle{named}
\bibliography{ijcai19}

\end{document}